\documentclass[runningheads]{llncs}
\usepackage[T1]{fontenc}
\usepackage{graphicx}
\usepackage{color}
\usepackage{hyperref}

\usepackage{booktabs}       
\usepackage{amsfonts}       
\usepackage{amsmath}
\usepackage{multirow}

\begin{document}
%
\title{AQBuc: Air Pollution Forecasting in Bucharest}
%
%
\author{Dragoș-Andrei Șerban\inst{1} \and
Răzvan-Alexandru Smădu\inst{1,2} \and
Dumitru-Clementin Cercel\inst{1,2}}
\authorrunning{D.-A. Șerban et al.}
%
\institute{NUST POLITEHNICA Bucharest, Bucharest, Romania \and
Academy of Romanian Scientists, Bucharest, Romania \\
\email{dumitru.cercel@upb.ro}
}
\maketitle              
\begin{abstract}
In recent years, air pollution, especially the particulate matter $2.5\,\mu\text{m}$ (PM2.5), has become a growing concern mainly in urban areas. Exposure to air pollution is linked to the development of numerous health conditions, such as the aggravation of respiratory diseases, cardiovascular disorders, impaired lung function, and even cancer or early death. Forecasting future levels of PM2.5 has become increasingly important, as it can provide early warnings and help prevent diseases. We propose AQBuc, a dataset for predicting PM2.5 pollution in Bucharest, Romania, introducing an understudied location, as most work in this field has focused on Asian cities. Our dataset allows us to benchmark models for predicting future levels of PM2.5 across various time horizons. We evaluate a broad range of methods, including machine learning models, such as linear regression and ensemble-based methods, deep learning models based on recurrent networks and transformers, and large language models (LLMs). Specialized time-series models handle temporal data better, especially on longer horizons, while tree ensembles are more cost-effective for short horizons. Meanwhile, LLMs struggle to capture continuous sensor dynamics without specialized architectural changes.

\end{abstract}

\section{Introduction}

Air pollution has become one of the most pressing issues of our time. Pollutants such as particulate matter $2.5\,\mu\text{m}$ (PM2.5) and $10\,\mu\text{m}$ (PM10) originate from sources including the combustion of gasoline, diesel, and wood. Due to their microscopic size, PM2.5 particles can penetrate deep into the lungs and enter the bloodstream, whereas larger PM10 particles are predominantly deposited in the upper airways. Even short-term exposures to high PM2.5 levels can cause serious problems such as chronic bronchitis, asthma attacks, and can be associated with premature mortality \cite{pm_and_health}.

According to the World Health Organization, the average annual levels of PM2.5 should not exceed $5\,\mu\text{g/m}^3$ \cite{WHO2021aqg}. Unfortunately, according to IQAir\footnote{\url{https://www.iqair.com/world-most-polluted-cities}, Accessed: 10 January 2026}, most of the metropolitan areas in the world exceed this threshold, with some of the highest levels recorded in 2024 in developing Asian cities such as Byrnihat and Delhi in India ($128.2\,\mu\text{g/m}^3$ and $108.3\,\mu\text{g/m}^3$, respectively) or Karaganda in Kazakhstan ($104.8\,\mu\text{g/m}^3$). Bucharest's average measured levels in 2024 were $15.7\,\mu\text{g/m}^3$, down from $20.3\,\mu\text{g/m}^3$ in 2018.
Forecasting PM2.5 levels allows governments, state agencies, and environmental organizations to anticipate higher probability periods of increased pollution, making decision-making easier during those periods and raising public awareness. Accurate PM2.5 forecast requires information on other environmental factors, such as air temperature, nitrogen dioxide ($\text{NO}_2$), sulfur dioxide ($\text{SO}_2$), ozone ($\text{O}_3$), and carbon monoxide (CO).

This paper focuses on evaluating various classical machine learning (ML) and deep learning models for forecasting PM2.5 over multiple horizons: 1, 2, and 4 hours ahead for most models, and even 8 hours ahead for recurrent neural networks (RNNs) and transformers specifically designed for time series analysis. We train statistical models and neural networks using historical air quality and meteorological measurements from the proposed Bucharest dataset called AQBuc. Although other studies present ML models for PM2.5 forecasting, most do not offer a detailed comparison of how different architectures perform.

Our work brings the following contributions: (i) We built and processed a dataset in Bucharest containing both pollution (i.e., PM2.5, PM10, $\text{NO}_2$, $\text{SO}_2$, CO, and $\text{O}_3$) and meteorological data (i.e., air temperature, humidity, wind speed, and direction).
(ii) We implemented and compared multiple machine learning and deep learning algorithms, including hybrid models and encoder-decoder architectures.
(iii) We analyzed multiple transformer models (Darts \cite{3586589.3586713}, Informer \cite{zhou2021informer}, and PatchTST \cite{nie2022time}), and compared the T5 model \cite{2020t5} with and without retrieval-augmented generation (RAG) \cite{rag}, as well as zero-shot Gemini 1.5 Flash \cite{geminiteam2024gemini15unlockingmultimodal}.
(iv) The source code and the dataset are publicly available\footnote{\url{https://github.com/DragosSerban/air-pollution-forecasting-in-bucharest}, Accessed: 10 January 2026}.

\section{Related Work}

In recent years, machine learning and deep learning techniques have been widely applied to air quality forecasting to capture complex relationships between pollutants and meteorological conditions.
Mohammadzadeh et al. \cite{mohammadzadeh2024spatiotemporal} proposed a hybrid architecture that combined a graph convolutional network (GCN) \cite{DBLP:conf/iclr/KipfW17} and an exogenous long short-term memory (LSTM) \cite{hochreiter1997long}.
The multivariate LSTM considered the PM2.5 concentration, along with temperature, humidity, pressure, wind speed, and wind direction. The PM2.5 and weather data were collected by several stations in Michigan, United States, from January 2015 to December 2019.
Bialka et al. \cite{zajkeckaautomated} focused on forecasting PM2.5 with a 6-hour horizon using three methods: multi-layer perceptron (MLP), LSTM, and echo state networks (ESN). The models were trained on historical datasets collected by stations near Krakow, Poland, containing PM2.5 values, exogenous variables, meteorological forecast data, and data from nearby stations. ESN achieved the best results, with LSTM and MLP lagging behind.
Cáceres-Tello and Galán-Hernández \cite{caceres2024analysis} employed a hybrid Prophet-LSTM model on daily measurements from January 2019 to June 2024 from seven different districts in Madrid. Prophet models seasonality, trends, and special events, such as holidays or significant known incidents, while the LSTM captures the residuals (e.g., noise and temporary episodes) by modeling short- and long-term dependencies.

Several works have addressed air pollution forecasting using data from China. Liu et al. \cite{liu2025deep} combined two self-attention mechanisms with LSTM to predict PM2.5.
The first attention layer learned from previous data, while the second one captured the relationships among the other measurements in the sequence.
Particle swarm optimization automatically adjusted the batch size and learning rate. The hybrid model was evaluated on datasets from Wuhan and Nanchang, outperforming the other baselines.
Peng et al. \cite{peng2022machine} evaluated extreme gradient boosting (XGBoost) \cite{xgboost} and a fully connected neural network on measurements from two stations in Hunan Province, central China, collected over 2021. Each sample included PM2.5 levels, wind speed and direction, air temperature, humidity, atmospheric pressure, and rainfall. The boosting method achieved higher performance. 
Bai et al. \cite{bai2019hourly} introduced a stacked autoencoder (SAE) that learned a compressed representation, while a fully connected layer predicted the final class. The SAE layer was trained unsupervised, while the dense layer was trained using supervision on data near Beijing.
Qin et al. \cite{qin2025sfdformer} presented a new transformer-based model, SFDformer, which combined time-series decomposition with the Fourier transform to capture trends and seasonality and used a sparse attention mechanism to identify the most relevant patterns. The model outperformed other transformer-based architectures.

Despite these recent advances, two critical gaps remain unresolved. First, most recent benchmarks focus heavily on East Asia or North America, leaving European environments largely underrepresented. Second, existing studies typically evaluate only a narrow subset of architectures, lacking a unified cross-paradigm comparison. AQBuc addresses these gaps by introducing a multi-year, multi-pollutant dataset from Bucharest and establishing an extensive empirical benchmark across these diverse model families.
 
\section{Dataset}

\subsection{Construction of AQBuc} 

\textbf{Dataset Collection.}
The corpus that we use for this work was downloaded from \url{www.calitateaer.ro}, a Romanian website managed by the National Agency for Environmental Protection. This agency is subordinate to the Ministry of Environment and has the authority to implement environmental policies. The website provides real-time data collected from numerous stations that measure air quality in Bucharest and throughout Romania.
We name our dataset AQBuc. It presents a report on historical pollution data measured by the B-1 RNMCA BUC station near Mill Lake (ro. ``Lacul Morii''), located in the 6th District of Bucharest. This station was chosen for its high levels of pollutants (i.e., PM2.5, PM10, $\text{NO}_2$, $\text{SO}_2$, CO, and $\text{O}_3$). In addition, the station measures atmospheric attributes, including temperature and humidity, as well as wind speed and direction.
We downloaded the Bucharest dataset as an Excel file containing 36,060 rows of hourly measurements from August 1, 2019 to July 31, 2023. It is similar in size to the dataset proposed by Mohammadzadeh et al. \cite{mohammadzadeh2024spatiotemporal}, which used four years of hourly data for training and testing.

\textbf{Dataset Format.} 
Each measurement record (see Table \ref{tab:detailed_stats}) from the dataset contains the following variables: a timestamp with the format \texttt{yyyy-mm-dd hh}, levels of $\text{NO}_2$, $\text{SO}_2$, CO, and $\text{O}_3$, wind speed and direction, air temperature, and concentrations of PM2.5 and PM10 air pollutants. The measurements in the Bucharest dataset contain relatively few missing data points and outliers, which we will focus on in the following sections.

\begin{table}[!t]
\caption{Summary statistics for 36,060 considered samples, including the minimum, maximum, average, and standard deviation for each variable.}
\label{tab:detailed_stats}
\scriptsize
\begin{tabular*}{\linewidth}{@{\extracolsep{\fill}}lcccc}
\toprule
\textbf{Variable (Unit)} & \textbf{Min} & \textbf{Max} & \textbf{Mean} & \textbf{Std} \\
\midrule
$\text{NO}_2$ (\(\mu\)g/m\textsuperscript{3}) & -4.68 & 178.50 & 26.48 & 19.19 \\
$\text{SO}_2$ (\(\mu\)g/m\textsuperscript{3}) & 0.67 & 314.12 & 5.14 & 3.00 \\
CO (mg/m\textsuperscript{3}) & -0.52 & 4.51 & 0.53 & 0.30 \\
$\text{O}_3$ (\(\mu\)g/m\textsuperscript{3}) & -4.18 & 187.74 & 44.56 & 29.55 \\
Wind direction (degrees) & 0.00 & 360.00 & 138.29 & 119.69 \\
Air temperature ($^\circ\text{C}$) & -7.86 & 40.40 & 14.19 & 9.35 \\
Wind speed (m/s) & 0.00 & 10.20 & 0.73 & 1.04 \\
PM10 (\(\mu\)g/m\textsuperscript{3}) & -9.72 & 627.36 & 25.62 & 19.79 \\
PM2.5 (\(\mu\)g/m\textsuperscript{3}) & -9.99 & 571.69 & 16.58 & 14.08 \\
\bottomrule
\end{tabular*}
\end{table}

\subsection{Data Processing}

\textbf{Dataset Cleaning.}
We select the columns containing numerical data (i.e., $\text{NO}_2$, $\text{SO}_2$, CO, $\text{O}_3$, PM10, and PM2.5) and convert the floating-point values from comma to dot format. The dataset also contains some missing values, mainly for $\text{SO}_2$, CO, and $\text{O}_3$.
We observe some outliers and negative values for the PM2.5 concentration. The concentration cannot be negative, as it represents a mass concentration of particulate matter. The measurements may contain errors due to faulty sensors or problematic correction methods.
Inspired by Bialka et al. \cite{zajkeckaautomated}, we address outliers by employing the forward-backward exponential weighted moving average (FBEWMA) algorithm \cite{hodge2004survey}. This method identifies outliers by detecting considerable deviations of current measurements from the trend of previous or subsequent moments. Therefore, we apply an exponentially weighted moving average to each numerical column in both temporal directions, then average the results to obtain the FBEWMA. If the chosen threshold (in our case, 5) is exceeded, the value is considered an outlier and replaced with NaN. In the end, we found 1,795 such values in the entire dataset.
After removing outliers, we use linear interpolation to approximate the missing values in the dataset. To address harmful pollutants, we applied corrections to $\text{NO}_2$, $\text{SO}_2$, CO, $\text{O}_3$, PM10, PM2.5, and set all negative values to 0.

\textbf{Data Normalization.}
For neural networks, tree ensembles, and regression models (i.e., LSTM, GRU \cite{cho2014learning}, linear regression, SVR, random forests \cite{random_forest}, XGBoost, Informer, PatchTST, and Darts), we applied min-max normalization, scaling values to $[0, 1]$. We did not scale features for models such as ARIMAX \cite{arimax}, SARIMAX \cite{sarimax}, and T5, as they work with the actual time-series values.

\textbf{Feature Engineering.}
We added new columns to improve the forecast for the PM2.5 values:
(i) Lag features for the PM2.5 pollutant were added for the models that lack intrinsic sequential mechanisms (i.e., regression models, MLPs, boosting, and bagging algorithms), to capture recent temporal trends and short-term dynamics. We added them to better understand recent trends and how this pollutant has changed over the past hours.
(ii) Lag features for the other pollutants (i.e., $\text{NO}_2$, $\text{SO}_2$, $\text{O}_3$, and CO).
(iii) Day of the week (Monday = 0 to Sunday = 6), capturing weekday dynamics; for example, traffic volume and industrial activity are typically higher on working days.
(iv) Weekend indicator (binary flag) suggesting whether the measurement was done at the end of the week or not, accounting for lower traffic and industrial emissions during weekends.
(v) Season (winter = 0, spring = 1, summer = 2, and autumn = 3), capturing seasonal dynamics such as increased fuel consumption for winter heating, resulting in higher levels of pollution.

\subsection{Data Analysis}

\begin{figure}[!ht]
    \centering
    \includegraphics[width=\linewidth]{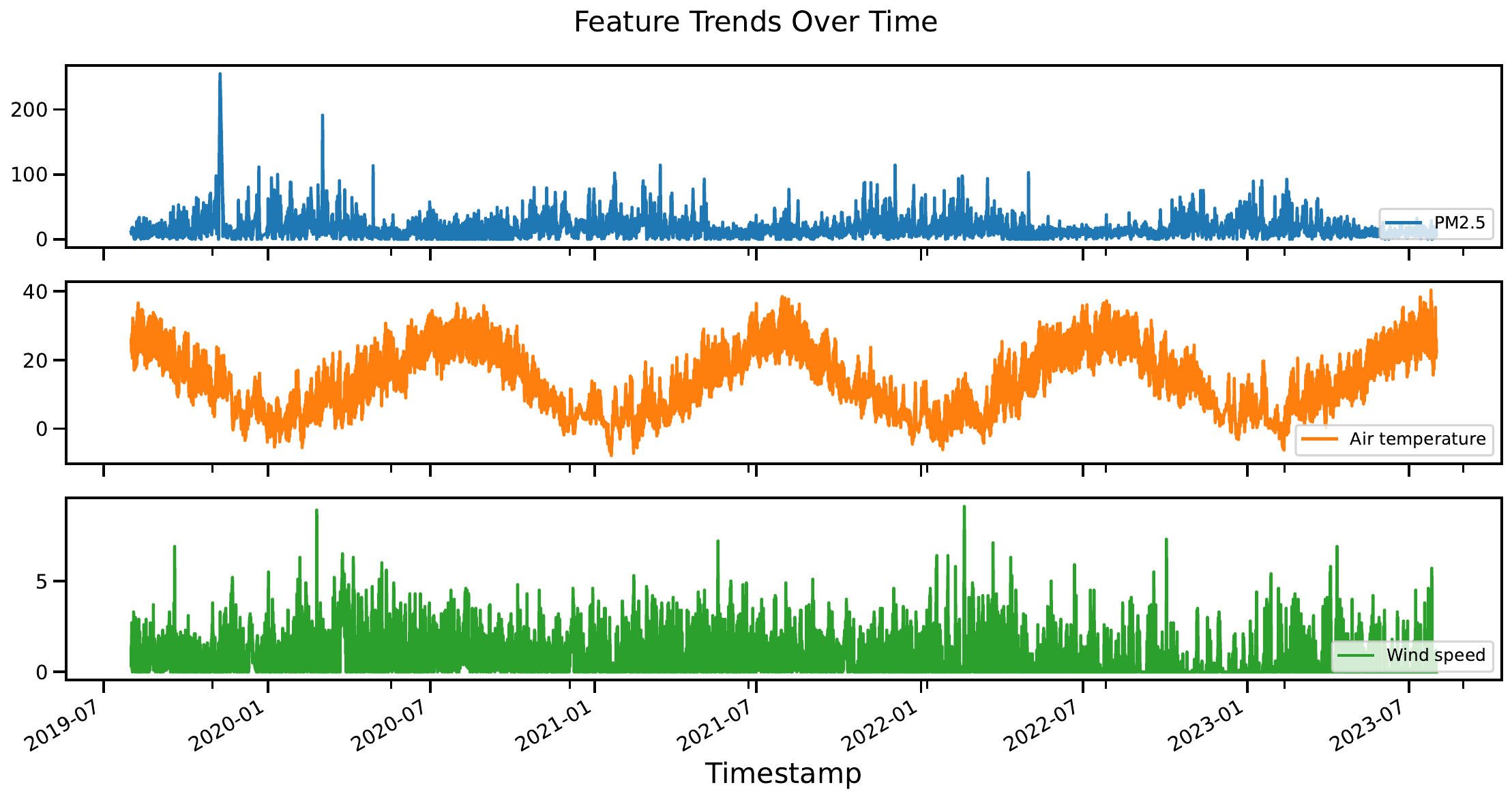}
    \caption{Evolution of PM2.5, air temperature, and wind speed between 2019 and 2023.}
    \label{fig:evolution_quality}
\end{figure}

\begin{figure}[!ht]
    \centering
    \includegraphics[width=\linewidth,trim={0 9.5cm 0 0},clip]{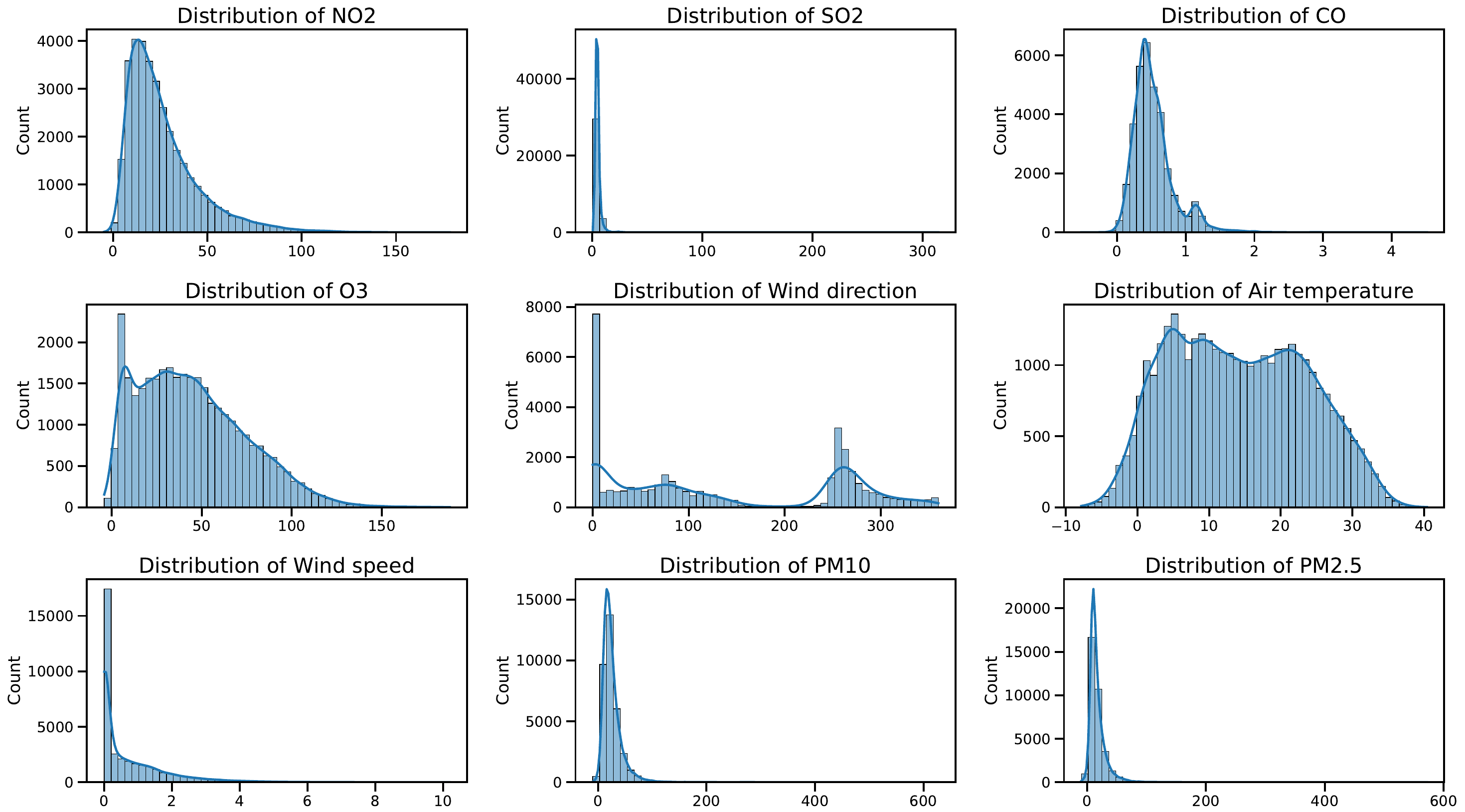}
    \caption{Distribution of hourly sensor values for all nine attributes (see Table \ref{tab:detailed_stats} for the physical units on the horizontal axis).}
    \label{fig:distribution_values}
\end{figure}

\textbf{Series Analysis.}
We did not identify any trend in the attributes. However, Figure \ref{fig:evolution_quality} shows a recurring yearly behavior in air temperature (higher in summer, lower in winter) and wind speed, along with seasonal winter peaks in PM2.5 concentrations. A daily pattern is also evident in the PM2.5 concentration column, with higher values during the day and lower values at night.
Attributes such as $\text{NO}_2$, $\text{SO}_2$, CO, $\text{O}_3$, wind speed, PM10, and PM2.5 exhibit a positively skewed (right-skewed) distribution across the four-year observation window (see Figure \ref{fig:distribution_values}). Typical readings reflect moderate urban baseline levels, while long right tails capture episodic pollution spikes, predominantly occurring during winter conditions. Therefore, the mean exceeds the median for these attributes.
The air temperature distribution is nearly symmetrical, with a flattened peak, suggesting a temperate continental climate in Romania.
Regarding the wind direction, a multimodal distribution is observed, with two main peaks: one at $0^\circ$ (north) and the other at $260^\circ$ (west). In addition, a range of values is observed for the south-southwest direction (between $160^\circ$ and $240^\circ$).

\textbf{Stationarity Analysis.}
To evaluate the stationarity of the time series, two tests have been performed: the Augmented Dickey-Fuller (ADF) test \cite{Dickey01061979}, which checks whether the time series is stationary or if it needs to be differenced to be stationary, and the Kwiatkowski-Phillips-Shin (KPSS) test \cite{KWIATKOWSKI1992159}, which verifies the stationarity around a mean or a trend. A variable is completely stationary if the null hypothesis is rejected by the ADF, but not by the KPSS \cite{zajkeckaautomated}.
The results shown in Table \ref{tab:stationary_statistics} indicate that the attributes, namely PM2.5, $\text{NO}_2$, $\text{SO}_2$, CO, $\text{O}_3$, the air temperature, wind speed, and wind direction, reject the null hypothesis in both tests. This means that they are stationary in differences, but not in trends. Furthermore, the PM10 pollutant is the only variable for which the null hypothesis is rejected by the ADF test but not by the KPSS test, suggesting that PM10 is stationary. These findings have direct implications for modeling: they confirm that statistical models like ARIMAX and SARIMAX require first-order differencing ($d = 1$) to avoid spurious regressions, while neural architectures benefit from feature normalization to stabilize learning across non-stationary seasonal shifts.

\begin{table}[!t]
\centering
\caption{Results of ADF and KPSS tests, including test statistics and 5\% critical values, for assessing stationarity of the time series attributes.}
\label{tab:stationary_statistics}
\scriptsize
\begin{tabular*}{\linewidth}{@{\extracolsep{\fill}} l c c c c }
\toprule
\multirow{2}{*}{\textbf{Attribute}} & \multicolumn{2}{c}{\textbf{ADF}} & \multicolumn{2}{c}{\textbf{KPSS}} \\
\cmidrule{2-3} \cmidrule{4-5}
 & \textbf{Test} & \textbf{5\% crit. val.} & \textbf{Test} & \textbf{5\% crit. val.} \\
\midrule
PM2.5 & -15.28 & -2.86 & 0.78 & 0.46 \\
PM10 & -15.53 & -2.86 & 0.36 & 0.46 \\
$\text{NO}_2$ & -16.17 & -2.86 & 2.37 & 0.46 \\
$\text{SO}_2$ & -12.24 & -2.86 & 3.80 & 0.46 \\
CO & -9.39 & -2.86 & 2.87 & 0.46 \\
$\text{O}_3$ & -7.99 & -2.86 & 0.68 & 0.46 \\
Air temperature  & -5.32 & -2.86 & 0.50 & 0.46 \\
Wind speed & -18.21 & -2.86 & 10.68 & 0.46 \\
Wind direction & -14.88 & -2.86 & 13.53 & 0.46 \\
\bottomrule
\end{tabular*}
\end{table}

\begin{figure}[!ht]
    \centering
    \includegraphics[width=0.65\linewidth]{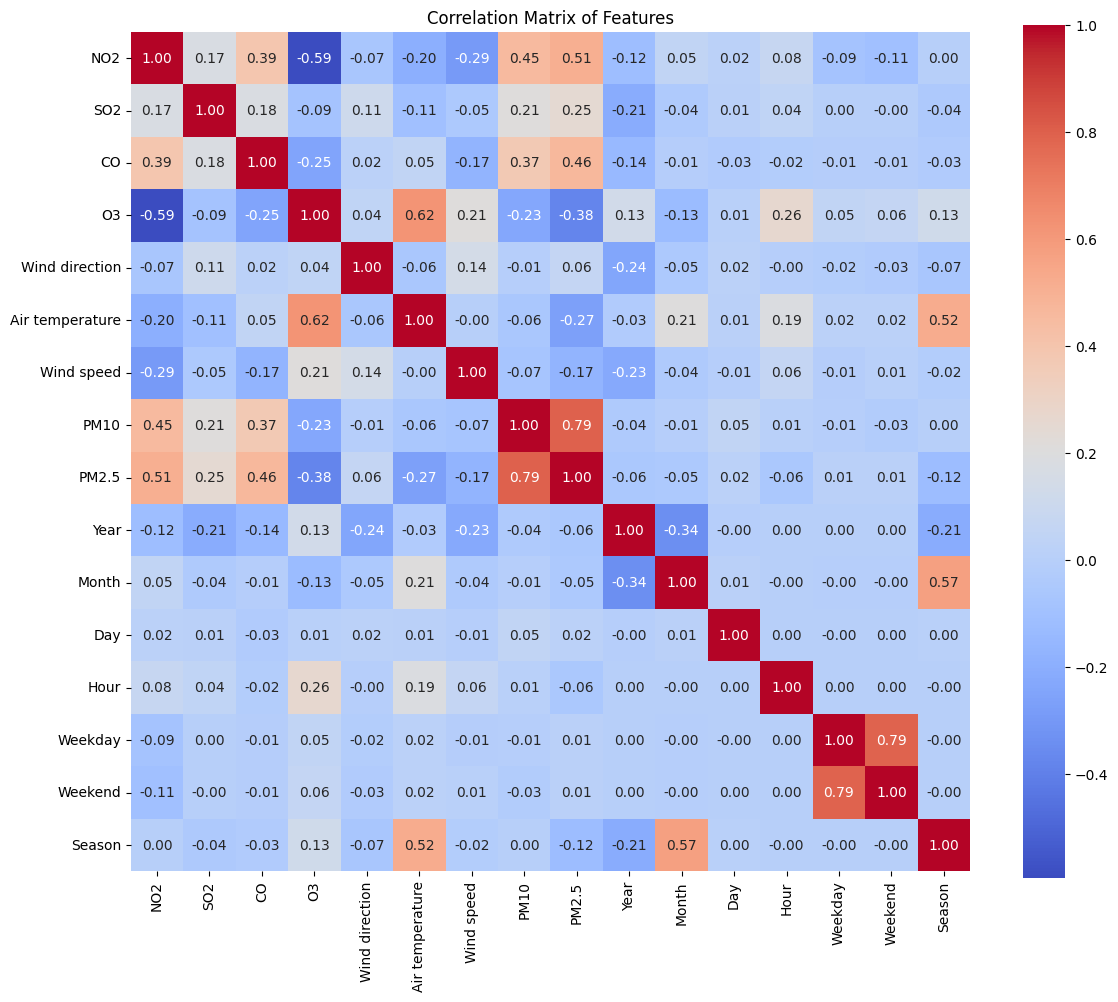}
    \caption{Correlation matrix between measured attributes.}
    \label{fig:correlation}
\end{figure}

\textbf{Correlation Analysis.}
For this analysis, the Pearson correlation coefficient was used to assess the linear relationship between numerical variables. Figure \ref{fig:correlation} illustrates the pairwise Pearson correlations between the measured attributes. Between PM2.5 and PM10, there is a strong positive correlation with a Pearson coefficient of 0.79. Between air temperature and season, a moderate positive correlation of 0.52 is observed, as shown in the evolution of the temperature variable by season. However, there is a moderately strong negative correlation between ozone and $\text{NO}_2$, with a value of -0.59, suggesting that higher levels of $\text{NO}_2$ are associated with lower ozone levels. The strong linear dependencies between PM2.5 and PM10, as well as the photochemical relationship between $\text{NO}_2$ and $\text{O}_3$, confirm that co-pollutants and weather conditions carry substantial mutual information, justifying a multivariate forecasting approach over univariate models.

\textbf{Data Splits.}
The dataset is chronologically split into a training set (consisting of 80\% of the samples, used to learn the relationships between the independent variables and the target attribute) and a held-out test set (comprising the last 20\% of samples, used for model performance evaluation).

\section{Method}

We evaluate several models, ranging from traditional machine learning models to neural networks, transformer-based models, and LLMs.

\textbf{Traditional Machine Learning Models.} We benchmark various machine learning algorithms, including linear regression, least absolute shrinkage and selection operator (Lasso) regression, Ridge regression, ElasticNet, and support vector regression (SVR), with the following hyperparameters.

For \textbf{Lasso regression}, we select the optimal regularization parameter $\alpha \in [0.0001, 10]$ via 4-fold cross-validation, to find the best value $\alpha=0.00687$. Before training, we apply MinMax scaling to the dataset to ensure that all features contribute equally to the result and to reduce bias.
For \textbf{Ridge regression}, we employ a 5-fold cross-validation method to search for the optimal alpha parameter over 50 values, logarithmically spaced within the previously mentioned interval. The best $\alpha$ found is 1.75751. The best model is selected by maximizing the negative mean squared error. In addition, MinMax scaling is applied to improve the convergence speed.
In \textbf{ElasticNet regression} experiments, we use 5-fold cross-validation to search for $\alpha$ logarithmically between 0.0001 and 10, and \texttt{l1\_ratio} between 0 and 1.
To select the optimal \textbf{SVR} parameters, we use a grid search with a 5-fold cross-validation. It searches for the best margin $C$ (0.1--100, optimal was 100) and kernel (linear and radial basis, optimal was linear) functions. The best model is the one with the lowest mean squared error.

Decision-tree-based methods are also considered through random forests and XGBoost. 
In \textbf{random forests}, the final prediction is the majority vote in classification tasks and the mean of all the trees' predictions in regression tasks. We tune the number of trees, depth, and split criteria using 3-fold cross-validation. The best hyperparameters are 200 estimators, at least 2 leaves, and a minimum branching factor of 2. Also, we find that the maximum depth is \texttt{None}, suggesting that the model captures more complex patterns in the data, but may be more prone to overfitting.
\textbf{XGBoost} is a gradient boosting method that trains several decision trees sequentially, and the final prediction is a weighted sum of all trees' predictions. We use 3-fold cross-validation. The number of trees is tested at 100 and 200, and the maximum tree depth at 3, 6, and 10. The learning rate is chosen from a list of values: 0.01, 0.1, and 0.2. The results show that the model is selected to reduce overfitting and increase performance. The maximum depth of 3 suggests the model prefers smaller trees, while the large number of estimators (200) helps create a more complex and powerful model.

Finally, we consider ARIMAX \cite{arimax} and SARIMAX \cite{sarimax}, two methods designed for time series analysis that include an autoregressive component (AR), a differencing component (I), and a moving-average component (MA). SARIMAX, in addition to ARIMAX, also accounts for seasonal cycles.
During \textbf{ARIMAX} fine-tuning, we search for the autoregressive parameter $p$ and the moving-average parameter $q$, while keeping the differencing parameter at 1. All possible natural configurations for $p$ and $q$ in the range of 0 to 2 are generated, and for each combination, the model is trained on exogenous variables. The optimal orders are $p = 1$ and $q = 1$. The model uses one lag value of the target variable to predict the current value and includes one lag of the forecast error in the prediction.
Since \textbf{SARIMAX} includes a seasonal component, which adds four additional hyperparameters to the ARIMAX model, a search was conducted to determine the optimal parameters ($p$, $d$, $q$) for the nonseasonal order and ($P$, $D$, $Q$, $s$) for the seasonal order. Since PM2.5 shows a clear 24-hour seasonality in urban environments, the parameter $s$ is set to 24. The other six hyperparameters are combined to evaluate which set of values has the best performance. We set $P=1$, $D=0$, and $Q=1$, while $p$ and $q$ are selected as 0 and 1.

\textbf{Neural Network Architectures.} We consider simple feed-forward networks based on \textbf{MLP} \cite{popescu2009multilayer}, and a more complex model based on \textbf{Kolmogorov-Arnold Networks (KANs)} \cite{kan_paper} to capture the local smoothness of the features, and only some of those that actually influence the result.
MLPs, KANs, and hybrid architectures between these two architectures (i.e., MLP-KAN and KAN-MLP) are tested.
All four models have input layers with the same number of features and output layers with 4 neurons to forecast PM2.5. The KAN consists of two hidden layers with $2n$ and $n/2$ neurons, where $n$ is the number of input features, while the MLP uses the same hidden sizes with ReLU activations and dropout for regularization. The MLP–KAN hybrid passes a fully connected layer ($2n$ neurons, ReLU) into a KAN block, and the KAN–MLP hybrid passes a KAN block into a fully connected layer, with spline functions used in the hidden layer of the KAN block.

\textbf{LSTM} and \textbf{gated recurrent units (GRUs)} are evaluated in bidirectional settings (i.e., BiLSTM, BiGRU), stacked layers, or combined with convolutional neural networks and encoder-decoder architectures \cite{cho2014learning}.
For both LSTM and GRU models, we evaluate PM2.5 forecasting performance on several architectures trained after MinMax scaling, using the mean squared error (MSE) loss and the Adam optimizer. We test single-layer, stacked, bidirectional, and sequence-to-sequence LSTMs, all with 64 units per layer, and output layers with 8 neurons to predict the next 8 hours. Dropout regularization and early stopping are applied to prevent overfitting. The sequence-to-sequence model uses an encoder-decoder structure to summarize the past 24 hours and generate multistep forecasts. The models converge in less than 100 epochs.

We train and test a \textbf{CNN} model and four hybrid architectures (i.e., LSTM-CNN, CNN-LSTM, CNN-GRU, and CNN-LSTM encoder-decoder) that leverage CNNs and RNNs to capture both spatial and temporal patterns in the dataset. Convolutional layers utilize 32--64 filters to extract local patterns, while recurrent layers employ 64--128 units per layer to capture sequential dependencies. We use the MSE loss function and the Adam optimizer for training. Dropout regularization and early stopping are applied to prevent overfitting.

In addition, we utilize the transformer architectures \cite{vaswani2017attention}, which replace recurrence with self-attention.
We test three transformers for PM2.5 forecasting: Darts, Informer, and PatchTST, predicting 1, 2, 4, and 8 hours from the last 48 measurements. 
The Darts transformer \cite{3586589.3586713} is based on the original transformer model and is fine-tuned for time series. 
The Informer model \cite{zhou2021informer} is designed to achieve good performance on long sequences.
PatchTST \cite{nie2022time} brings features of patching and channel independence.
The Darts transformer employs a three-layer encoder-decoder architecture with multi-head attention, feed-forward layers of 128 neurons, and an embedding dimension of \texttt{d\_model=64}. The Informer features a two-layer encoder with ProbSparse self-attention (4 heads) and feed-forward layers of 128 neurons, as well as a single-layer decoder incorporating both cross- and self-attention. PatchTST is encoder-only with two layers of multi-head attention (4 heads) and feed-forward layers, embedding patches of 8 steps into a 128-dimensional space. All models use dropout for regularization, linear output layers for 8-step predictions, and early stopping to prevent overfitting.

\textbf{Large Language Models.} Motivated by recent studies \cite{roitero2025leveraging}, we evaluate whether general-purpose LLMs can effectively forecast continuous air pollution conditions. In forecasting, a new dataset is used to fine-tune the model, enabling it to better understand temporal dependencies.
We employ T5-small (60M parameters, 6 layers, 8 attention heads) and T5-base (220M parameters, 12 layers, 12 attention heads) models. All models are trained for 5 epochs using the Adam optimizer, a batch size of 8, a learning rate of 3e-4, and early stopping.
Since LLMs process text, the inputs are reformulated as natural-language prompts that contain feature values and recent PM2.5 measurements, with the outputs as target PM2.5 values.
Retrieval-Augmented Generation (RAG) \cite{rag} enables the language model to condition its predictions on semantically similar historical patterns retrieved from an external database. Similar to Roitero et al. \cite{roitero2025leveraging}, we combine the T5 language model with RAG to improve air pollution forecasting, where each input is encoded using paraphrase-MiniLM-L6-v2 and stored in a FAISS\footnote{\url{https://github.com/facebookresearch/faiss}, Accessed: 10 January 2026} database. During inference, the two most similar past samples are retrieved and appended to the input to improve forecasting accuracy.
We also evaluate the Gemini 1.5 Flash model \cite{geminiteam2024gemini15unlockingmultimodal} in the zero-shot prompting strategy. The prompt template for querying the model is defined below:

\begin{quote}
    ``Given the past \texttt{<history\_window>} hourly PM2.5 measurements: \\
    \texttt{<history\_series>} \\
    Predict the next \texttt{<forecast\_horizon>} PM2.5 values. Output the result strictly as a comma-separated list of floating-point numbers enclosed in brackets (e.g., [10.1, 10.2]). Do not provide any code, markdown formatting, or explanatory text.''
\end{quote}

\section{Results}

The performance of the models is evaluated using the mean absolute error (MAE), the root mean squared error (RMSE), and the coefficient of determination ($R^2$) \cite{hyndman2014forecasting}. These metrics were selected because MAE captures the average physical error without outlier distortion, RMSE penalizes large pollution spikes, and $R^2$ measures temporal variance across horizons.

\begin{table*}[!ht]
\caption{Results across various algorithms forecasting horizons of 1, 2, and 4 hours.}
\label{tab:results}
\centering
\scriptsize
\begin{tabular}{lccccccccc}
\toprule
 \textbf{Horizon}& \multicolumn{3}{c}{\textbf{1h}}& \multicolumn{3}{c}{\textbf{2h}}& \multicolumn{3}{c}{\textbf{4h}} \\
\cmidrule{2-4} \cmidrule{5-7} \cmidrule{8-10}
\textbf{Method}& \textbf{MAE}& \textbf{RMSE}& \textbf{R\textsuperscript{2}}& \textbf{MAE}& \textbf{RMSE}& \textbf{R\textsuperscript{2}}& \textbf{MAE}& \textbf{RMSE}& \textbf{R\textsuperscript{2}}\\
\midrule
 \multicolumn{10}{c}{Linear Regression Models}\\
\midrule
Linear Regression& 4.79& 6.55& 0.716& 5.35& 7.53& 0.625& 6.18& 9.03&0.462\\
Lasso Regression& 4.75& 6.52& 0.719& 5.30& 7.50& 0.628& 6.15& 9.00&0.464\\
Ridge Regression& 4.70& 6.25& 0.742& 5.24& 7.21& 0.657& 6.05& 8.75&0.493\\
ElasticNet& 4.80& 6.57& 0.714& 5.38& 7.60& 0.618& 6.18& 9.04&0.460\\
\midrule
 \multicolumn{10}{c}{Classical Machine Learning Models}\\
\midrule
SVR& 4.25& 5.82& 0.776& 4.83& 6.88& 0.687& 5.75& 8.57&0.514\\
Random Forest& 3.50& 5.26& 0.817& 4.28& 6.51& 0.720& 5.38& 8.20&0.556\\
XGBoost& 3.48& 5.21& 0.820& 4.23& 6.45& 0.725& 5.25& 8.06&0.571\\
\midrule
 \multicolumn{10}{c}{Statistical Time Series Models}\\
\midrule
ARIMAX& 4.52& 5.69& 0.786& 5.11& 6.58& 0.714& 6.41& 8.74&0.495\\
SARIMAX& 4.47& 5.68& 0.787& 5.04& 6.55& 0.716& 6.36& 8.73&0.497\\
\midrule
 \multicolumn{10}{c}{Feed-forward Networks}\\
\midrule
 MLP& 4.24& 5.72& 0.783& 5.32& 7.59& 0.619& 5.51& 8.34&0.540\\
 KAN& 4.36& 5.88& 0.771& 5.44& 7.44& 0.634& 6.34& 9.15&0.446\\
 MLP-KAN& 4.24& 5.90& 0.770& 5.57& 7.64& 0.614& 6.03& 8.87&0.480\\
 KAN-MLP& 4.01& 5.32& 0.812& 4.86& 6.70& 0.703& 5.60& 8.29&0.545\\
\midrule
 \multicolumn{10}{c}{Recurrent Neural Networks}\\
\midrule
 RNN& 3.79& 5.29& 0.815& 4.38& 6.36& 0.733& 5.41& 8.34&0.540\\
 LSTM& 3.72& 5.21& 0.821& 4.05& 6.04& 0.759& 5.18& 8.05&0.572\\
 BiLSTM& 3.93& 5.51& 0.799& 4.46& 6.59& 0.713& 5.52& 8.53&0.519\\
 Three-layer LSTM& 3.37& 4.90& 0.841& 4.19& 6.23& 0.744& 5.29& 8.07&0.570\\
 Three-layer BiLSTM& 3.54& 5.13& 0.826& 4.30& 6.47& 0.723& 5.41& 8.36&0.538\\
 LSTM-ED& \textit{3.16}& \textbf{4.50}& \textbf{0.866}& 4.08& 6.02& 0.760& 5.31& 8.00&0.577\\
 GRU& 3.78& 5.14& 0.825& 4.36& 6.25& 0.742& 5.39& 8.24&0.551\\
 BiGRU& 3.83& 5.40& 0.807& 4.38& 6.41& 0.728& 5.46& 8.11&0.565\\
 Three-layer GRU& 3.60& 4.96& 0.837& 4.40& 6.27& 0.740& 5.27& 7.93&0.585\\
 Three-layer BiGRU& 3.32& 4.67& 0.856& 4.07& \textit{5.95}& \textit{0.766}& 4.93& \textit{7.57}& 0.622\\
 GRU-ED& 3.46& 4.84& 0.845& 4.22& 6.15& 0.750& 5.29& 7.75&0.603\\
\midrule
 \multicolumn{10}{c}{CNN and Hybrid Architectures}\\
\midrule
 CNN& 4.64& 6.57& 0.715& 5.01& 7.31& 0.647& 5.72& 8.57&0.515\\
 LSTM-CNN& 4.39& 6.36& 0.732& 4.81& 7.04& 0.668& 5.42& 8.22&0.553\\
 CNN-LSTM& 3.63& 5.17& 0.823& 4.24& 6.19& 0.747& 5.17& 7.72&0.606\\
 CNN-LSTM-ED& 3.69& 5.41& 0.806& 4.21& 6.29& 0.739& 5.07& 7.74&0.604\\
 CNN-GRU& 3.70& 5.16& 0.824& 4.28& 6.13& 0.752& 5.08& 7.58& 0.621\\
\midrule
 \multicolumn{10}{c}{Transformers for Time Series}\\
\midrule
 Transformer (Darts)& 3.54& 5.09& 0.831& \textit{4.02}& 6.15& 0.753& \textit{4.85}& \textit{7.57}&\textit{0.626}\\
 Informer& 3.28& 4.74& 0.851& \textbf{3.87}& \textbf{5.80}& \textbf{0.776}& \textbf{4.81}& \textbf{7.45}&\textbf{0.631}\\
 PatchTST& \textbf{3.06}& \textit{4.53}& \textit{0.864}& \textit{4.02}& 6.15& 0.750& 5.08& 7.83&0.594\\
\midrule
 \multicolumn{10}{c}{Pre-trained Language Models}\\
\midrule
 T5-small& 3.93& 6.11& 0.785& 5.11& 8.09& 0.623& 7.17& 11.43& 0.246\\
 T5-small with RAG& 3.91& 6.16& 0.749& 5.24& 8.48& 0.524& 6.88& 11.16& 0.175\\
 T5-base& 4.26& 6.50& 0.715& 5.12& 8.06& 0.530& 6.16& 10.05&0.331\\
 T5-base with RAG& 4.32& 6.51& 0.788& 4.84& 8.01& 0.575& 6.14& 10.08&0.327\\
\midrule
 \multicolumn{10}{c}{Prompt Engineering}\\
\midrule
 Gemini 1.5 Flash& 4.03& 5.66& 0.787& 6.06& 8.50& 0.518& 8.12& 11.63&0.132\\
\bottomrule
\end{tabular}
\end{table*}

As shown in Table \ref{tab:results}, Random Forest and XGBoost exhibit the highest performance among classical machine learning models. They combine multiple learners, which enables them to capture more complex patterns in the data. Random Forest and XGBoost yield good results for PM2.5 prediction, with MAE values of approximately 3.50 and an $R^2$ score exceeding 0.80 for 1-hour horizons. SVR outperforms classical linear models, while ARIMAX and SARIMAX have better explanatory power than SVR. Linear regression models are the least effective among the tested methods, suggesting that they may struggle to capture complex information. Ridge regression performs slightly better due to $L_2$ regularization. For longer horizons, the ensemble methods maintain their lead.

Among deep learning models, most perform better than classical models, especially on longer horizons. Minor improvements over classical linear baselines are observed in the MLP, KAN, and hybrid architectures that combine the MLP and KAN layers. Higher performance is achieved by combining KAN layers with an MLP layer at the end (KAN-MLP), which approaches ensemble methods and outperforms regression and ARIMA models on all horizons. The hybrid KAN-MLP outperforms MLP-KAN, suggesting that capturing local patterns first and then feeding them into MLP layers is more effective.

RNNs provide further improvements. The bidirectional stacked GRU model performs best in this group because processing sequences in both directions helps to capture more dependencies. With stacked LSTMs, the unidirectional model performs better. Stacked RNNs generally outperform single-layer architectures because they learn richer temporal patterns. LSTM encoder-decoders show strong performance, explaining 86\% of the variability for 1-hour forecasts. Standard RNNs perform worse, struggling with longer sequences. CNN-RNN hybrids exhibit variable performance; CNN-LSTM outperforms LSTM-CNN, suggesting that extracting local feature patterns before recurrent processing enhances forecasts. GRU-based architectures often outperform their LSTM counterparts because they have a simpler structure.

Transformers for time-series forecasting achieve the best overall results. Informer performs best for longer horizons, while PatchTST outperforms the others for short-term predictions.
PatchTST achieves a 12.1\% reduction in MAE over XGBoost and a 3.2\% reduction over LSTM-ED at 1-hour ahead. At 4 hours, Informer achieves an 8.4\% MAE reduction over XGBoost and 22.2\% over linear regression. Furthermore, the performance ranking of Informer and PatchTST remains consistent across all three evaluation metrics, confirming their superior capability. 
Transformers designed for long-range dependencies, such as Informer, which uses ProbSparse attention, are especially effective for longer horizons, while patch-based tokenization helps short-term forecasts. The Darts transformer also performs well over all horizons, ranking second for 4-hour forecasts. In general, transformers outperform RNNs by capturing both long- and short-term dependencies without recurrence.
After fine-tuning, the T5 models perform decently over short-term horizons, but they are still outperformed by other deep learning models over longer horizons. The limited improvement from RAG can be attributed to the nature of semantic retrieval: retrieving past examples identifies similar numeric ranges from different days, but disrupts the exact continuous temporal alignment required for trajectory forecasting, introducing out-of-context noise.

\textbf{Limitations.} 
For air pollution forecasting, the dataset contains a single air quality measurement station, which may not fully capture the general trends and seasonal patterns in the Bucharest metropolitan area. Although the dataset includes meteorological and pollutant data, it lacks additional features such as real-time traffic data or industrial activity data that could improve predictions. Consequently, while the architectural comparison and temporal dynamics offer generalizable insights, the predictions cannot capture variations across other city districts without multi-station data. Due to computational constraints, training or fine-tuning some transformer models took considerable time, and larger models like T5-large could not be fine-tuned due to GPU memory limitations.

\section{Conclusions and Future Work}

This paper evaluated different models for air pollution forecasting in Bucharest, including classical machine learning (regression, SVR, random forest, XGBoost, ARIMA), neural architectures (MLPs, KANs, RNNs, CNN hybrids), transformers, and LLMs with RAG.
We prepared the dataset, added lag variables, removed outliers, imputed missing values, and analyzed attribute correlations. 
Our empirical results reveal several takeaways. First, specialized time-series transformers consistently outperform other models, especially on longer horizons. Second, advanced RNN-based architectures remain competitive for short-term 1-hour predictions. Third, tree-based ensemble models also yielded promising results on short-term predictions, demonstrating that aggregating predictions across multiple estimators can enhance predictions, while keeping computational overhead low. Finally, LLM and RAG-based methods struggle with time-series forecasting without making dedicated architectural changes.
Using data from multiple AQ stations offers an opportunity to develop and evaluate more advanced hybrid architectures, such as those that combine GCN with RNNs or transformers.

\subsubsection*{\ackname}
This work was supported by the NUST POLITEHNICA Bucharest through the PubArt program and the grant of the Academy of Romanian Scientists, AOȘR-TEAMS-IV Edition 2026-2027 “Digital Transformation in Science”.

\subsubsection*{Disclosure of Interests.}
The authors have no competing interests to declare that are relevant to the content of this article.


\bibliographystyle{splncs04}
\bibliography{mybibliography}

\end{document}